\documentclass{article}
\usepackage{iclr2026_conference,times}

\usepackage{amsmath,amssymb}
\usepackage{booktabs}
\usepackage{multirow}
\usepackage{graphicx}
\graphicspath{{./}{./figures/}}
\usepackage{subcaption}
\usepackage{hyperref}
\usepackage{url}
\usepackage{xcolor}
\usepackage{tcolorbox}
\usepackage{listings}
\usepackage{float}    
\usepackage{placeins} 
\tcbuselibrary{breakable}

\definecolor{okblue}{HTML}{0072B2}
\definecolor{okorange}{HTML}{E69F00}
\definecolor{okgreen}{HTML}{009E73}
\definecolor{okred}{HTML}{D55E00}

\iclrfinalcopy

\title{Can Blindfolded LLMs Still Trade? \\ An Anonymization-First Framework for Portfolio Optimization}

\author{Joohyoung Jeon \\
Korea University \\
Mirae Asset Securities \\
\AND
Hongchul Lee \\
Korea University \\
}

\begin{document}

\maketitle
\renewcommand{\thefootnote}{\fnsymbol{footnote}}
\footnotetext[1]{The findings and opinions expressed in this paper are those of the authors and do not represent the views of their affiliated institutions. Any remaining errors are the sole responsibility of the authors.}
\renewcommand{\thefootnote}{\arabic{footnote}}

\begin{abstract}
For LLM trading agents to be genuinely trustworthy, they must demonstrate understanding of market dynamics rather than exploitation of memorized ticker associations. Building responsible multi-agent systems demands rigorous signal validation: proving that predictions reflect legitimate patterns, not pre-trained recall. We address two sources of spurious performance: \textit{memorization bias} from ticker-specific pre-training, and \textit{survivorship bias} from flawed backtesting. Our approach is to blindfold the agents---anonymizing all identifiers---and verify whether meaningful signals persist. 

BlindTrade anonymizes tickers and company names, and four LLM agents output scores along with reasoning. We construct a GNN graph from reasoning embeddings and trade using PPO-DSR policy. On 2025 YTD (through 2025-08-01), we achieved Sharpe $1.40 \pm 0.22$ across 20 seeds and validated signal legitimacy through negative control experiments. To assess robustness beyond a single OOS window, we additionally evaluate an extended period (2024--2025), revealing market-regime dependency: the policy excels in volatile conditions but shows reduced alpha in trending bull markets.
\end{abstract}

\section{Introduction}
\label{sec:intro}
LLMs are increasingly applied to financial trading \citep{RockAlpha2025}, yet \citet{HLee2025} revealed that LLMs exhibit pre-trained preferences for technology stocks and large-cap stocks, leading to confirmation bias. This raises a fundamental question: are LLMs genuinely understanding markets, or merely memorizing patterns from training data containing phrases like ``Tesla stock surges''?

The backtesting environment compounds this problem. Survivorship bias (failed companies disappear from data, artificially inflating performance) and lookahead bias (treating tomorrow's news as already known) produce results that fail in real deployment.

To address these challenges, we propose BlindTrade, an anonymization-first LLM-GNN-RL framework. We anonymize S\&P 500 constituents daily, four specialized LLM agents evaluate stocks from different perspectives, results are aggregated through a GNN, and an RL policy determines the final portfolio.

\textbf{Contributions.}
(i) We introduce an \textbf{anonymization protocol} that replaces tickers and company-specific information (``AAPL'' $\rightarrow$ ``STOCK\_0026'') to block memorization.
(ii) We design a \textbf{specialized multi-agent system} where four agents (Momentum, News-Event, Mean-Reversion, Risk-Regime) evaluate stocks independently and output reasoning explaining their assessment.
(iii) We propose a \textbf{Semantic Graph Encoder (SemGAT)} that constructs a graph using sector connections and reasoning embedding similarity, enabling inter-stock relationship learning under anonymization.
(iv) We perform rigorous \textbf{signal validation} through IC analysis and negative control experiments (random shuffling), verifying that LLM signals have real predictive power and are not leakage artifacts.

\section{Related Work}
\label{sec:related}

\paragraph{LLM Trading Agents.}
Recently, attempts to directly use LLMs for trading have increased rapidly. FinGPT \citep{Yang2023} provides LoRA-based lightweight adaptation. FinMem \citep{Yu2023} adds hierarchical memory for past market patterns. TradingAgents \citep{Xiao2024} simulates a trading firm with multi-agent debates for decision-making.

However, real-time evaluations reveal limitations. LiveTradeBench \citep{HYu2025} evaluated LLM agents in real-time environments, and models that excelled on static benchmarks actually performed worse in actual trading. AI-Trader \citep{Fan2025} reaches a similar conclusion. In tests spanning U.S. stocks, Chinese A-shares, and cryptocurrencies, agents without risk management perform poorly in practice.

These studies suggest that LLMs can aid trading, yet it remains unclear \textbf{why} they are effective. They do not distinguish whether the models learned real market patterns or just memorized from training data. We address this problem through anonymization.

\paragraph{Limitations of Financial LLMs.}
FinBERT \citep{Araci2019} and BloombergGPT \citep{Wu2023} show tickers directly during training/evaluation, making it impossible to distinguish memorization from true understanding. Recent surveys \citep{LopezLira2023, Fu2025} raise the same concern.

\paragraph{Survivorship Bias and Lookahead Bias.}
Backtesting results are often vulnerable to survivorship and lookahead biases \citep{Elton1996, Bailey2014}. We address this by using only actual S\&P 500 constituents at each point in time.

\paragraph{GNNs for Finance.}
GNNs are effective for learning stock relationships \citep{Thakkar2021, Feng2019}, but most use fixed industry classification graphs. We dynamically construct edges using semantic similarity of LLM reasoning embeddings, allowing us to learn relationships even in an anonymized state.

\paragraph{Portfolio RL.}
\citet{Jiang2017} introduced end-to-end policy gradients, and \citet{Zhang2020} proposed regime-based approaches. However, most RL policies are black boxes. We explicitly expose intent variables (defensive/neutral/aggressive) for interpretability. Counter-intuitively, our Defensive mode shows higher turnover (2.9\%/day) because, without cash allocation, the policy diversifies across more stocks within the S\&P 500 universe, requiring frequent rebalancing. Aggressive mode concentrates on high-conviction positions and holds them for longer-term gains, resulting in minimal turnover (0.4\%/day). Neutral mode shows moderate activity (1.8\%/day). See Figure~\ref{fig:intent_metrics} for detailed intent-conditioned behavior analysis.

\section{Methodology}
\label{sec:methodology}

\subsection{Overall Pipeline}
BlindTrade consists of 6 stages (see Figure~\ref{fig:pipeline} in Appendix):

\begin{enumerate}
    \item \textbf{Data anonymization}: We collect S\&P 500 constituents daily and replace all tickers/company names/subsidiary names/product names with anonymous identifiers.
    \item \textbf{LLM feature generation}: Four specialized agents (Momentum, News-Event, Risk-Regime, Mean-Reversion) assign scores to each stock daily and output reasoning explaining ``why they made their assessment.''
    \item \textbf{IC validation}: We first check whether LLM outputs have real predictive power. We compare RAW variables and LLM variables using Spearman rank IC.
    \item \textbf{SemGAT encoding}: We learn inter-stock relationships as a graph. Same-sector stocks are connected, and additional edges are formed based on reasoning embedding similarity.
    \item \textbf{RL policy}: PPO determines portfolio weights. The policy internally decides risk posture (defensive/neutral/aggressive).
    \item \textbf{Backtest}: We evaluate OOS performance with 10bps transaction costs.
\end{enumerate}


\subsection{Data and Anonymization}
\label{sec:data}
We collect S\&P 500 constituents \textbf{point-in-time}. The stocks in S\&P 500 on January 2, 2020 differ from those on August 1, 2025. Index constituents change over time as companies are delisted or added. Ignoring this creates survivorship bias. We use only stocks that were actually S\&P 500 constituents on each date. Constituent information for each period is obtained via the EODHD API \citep{EODHD2024}. The total period spans 5.5 years (2020-01-02 to 2025-08-01), covering 1,403 trading days.

We map tickers to synthetic identifiers (e.g., ``AAPL'' $\rightarrow$ ``STOCK\_0026''), and proper nouns like ``Apple'', ``iPhone'', ``Tim Cook'' in news are replaced using Google Knowledge Graph API. We do not claim this blocks all leakage, but it at least blocks the path where LLM sees a ticker and decides ``it's Apple, so buy.''

\subsection{Multi-Agent LLM Feature Generation}
\label{sec:llm_features}

Four LLM agents evaluate stocks from different perspectives:
(i) \textbf{Momentum agent} checks whether price trends are strong and whether volume supports them;
(ii) \textbf{News-Event agent} reads anonymized news headlines and judges positive/negative sentiment;
(iii) \textbf{Mean-Reversion agent} finds overbought/oversold conditions---if a stock rose too much, it may fall; if it fell too much, it may bounce;
(iv) \textbf{Risk-Regime agent} looks at the overall market situation and judges whether systemic risk is high or low.

All agents operate under a strict knowledge cutoff, observing only 60 business days of data prior to time $t$ (from $t$-60 to $t$-1). Each agent receives a structured system prompt enforcing temporal constraints and deterministic JSON output (see Appendix~\ref{app:prompts} for full prompts). There is no lookahead bias from seeing tomorrow's news today. For news data, all headlines are anonymized, and news from t-60 to t-1 relative to time t is used. To avoid excessive context length, we limit input to 5 headlines per stock.

Importantly, we require LLM agents to produce explicit reasoning. For graph construction, we build a per-stock reasoning text by concatenating selected agents' reasoning snippets, then embed it into a 384-dimensional vector (implementation details in Appendix~\ref{app:impl}). We combine this embedding with numerical scores (7) and categorical states (3), forming a total 394-dimensional feature vector (see Tables~\ref{tab:feat_momentum}--\ref{tab:feat_risk} in Appendix for details).

\subsection{IC Validation}
\label{sec:validation}

We check whether LLM scores have real predictive power. We calculate IC (Information Coefficient, Spearman rank correlation) to see how correlated they are with returns 21 days later. We use h=21 days because medium-term signals are more stable and less noisy than daily correlations, providing a robust validation criterion.

Table~\ref{tab:ic_comparison} reports both the absolute IC and $\Delta$IC (RAW $\rightarrow$ LLM). We emphasize that a positive $\Delta$IC can reflect either (i) additional predictive signal, or (ii) removal of misleading inverse correlation by moving IC toward zero. In our experiments, the News-Event and Risk-Regime agents show statistically significant positive IC, while Momentum and Mean-Reversion primarily reduce misleading RAW correlations toward near-zero. The Risk-Regime agent showed the strongest $\Delta$IC (+0.017), with LLM IC = +0.011 (p $<$ 0.0001). On the 2025 holdout period specifically, Risk-Regime IC rose to 0.0515 (p $<$ 0.0001), confirming signal persistence out-of-sample.

\textbf{Feature selection via IC.} We use IC analysis as a screening criterion for LLM-derived features: we retain only features that remain informative on holdout (positive IC or positive $\Delta$IC). Each agent produces a richer structured output, but we include only the selected subset (marked ``Used'' in Tables~\ref{tab:feat_momentum}--\ref{tab:feat_risk}). Features evaluated at h=21 were verified to show positive (though weaker) IC at shorter horizons (h=1, h=5). While the absolute IC values are weak, there was clear improvement compared to using only RAW features.

\textbf{Input Features.} Each agent's input features were selected based on domain relevance (e.g., RSI for Mean-Reversion, beta for Risk-Regime). Output features are included if they showed positive $\Delta$IC or provided interpretable structure for downstream tasks. The Momentum agent receives price trends and volume indicators. The Risk-Regime agent receives beta, volatility, and tail risk. The Mean-Reversion agent receives RSI and Bollinger Band positions. The News-Event agent receives anonymized headlines with sentiment indicators (see Tables~\ref{tab:feat_momentum}--\ref{tab:feat_risk} for full specifications).

\begin{table}[t]
    \centering
    \caption{LLM vs RAW Feature IC Comparison (h=21 days). Negative RAW IC indicates inverse correlation; $\Delta$IC toward zero can remove misleading signals rather than adding predictive power. We therefore report both absolute IC and $\Delta$IC.}
    \label{tab:ic_comparison}
    \small
    \begin{tabular}{lccc}
        \toprule
        Agent & RAW IC ($p$) & LLM IC ($p$) & $\Delta$IC \\
        \midrule
        Momentum agent & $-0.019$ ($5{\times}10^{-5}$) & $+0.001$ (0.58) & $+0.020$ \\
        News-Event agent & $+0.003$ (0.12) & $+0.006$ ($3{\times}10^{-4}$) & $+0.003$ \\
        Mean-Reversion agent & $-0.005$ (0.26) & $-0.000$ (0.97) & $+0.005$ \\
        Risk-Regime agent & $-0.006$ (0.27) & $+0.011$ ($1{\times}10^{-4}$) & $+0.017$ \\
        \midrule
        \textbf{Average} & $-0.007$ & $+0.005$ & $+0.011$ \\
        \bottomrule
    \end{tabular}
\end{table}

\subsection{Semantic Graph Encoder (SemGAT)}
\label{sec:gnn}

Even with anonymization, sector information is preserved from point-in-time S\&P 500 constituent data via the EODHD API. So we can connect edges based on sector information of anonymized stock codes. However, in this experiment, we apply a Semantic Rewiring technique based on LLM reasoning. We use two types of edge connections: (i) \textbf{Sector edges} fully connect stocks in the same sector, and (ii) \textbf{Semantic edges} connect stocks if cosine similarity of reasoning embedding vectors exceeds 0.75 (top 10 neighbors per node). This allows the model to learn ``these two stocks were evaluated with similar reasoning'' even under anonymization.

The GNN structure is a 2-layer GATv2 encoder. Each stock is represented as a 394-dimensional feature vector (LLM scores + reasoning embedding vectors), which is transformed into a 128-dimensional node embedding. A distribution head predicts return distributions using HL-Gauss, and training combines distribution loss with pairwise ranking loss.

\textbf{Distributional Prediction.} Point estimation of next-day returns creates two problems: overfitting to extreme returns and inability to express uncertainty. We use HL-Gauss distribution prediction \citep{Bellemare2017} with 101 bins for next-day return ranking. Training combines distribution loss, pairwise ranking loss, market risk prediction, and Jensen-Shannon regularization.

\subsection{RL Policy (PPO-DSR)}
\label{sec:rl}

Once the GNN provides stock-level scores, the RL policy determines final portfolio weights. Our action space allocates weights only within the equity universe (no cash), so risk control is expressed via diversification and turnover. The policy architecture has three components (see Figure~\ref{fig:rl_policy_arch} in Appendix):

(i) \textbf{Intent Head (consensus-based)}: We aggregate agent outputs across all stocks into 4 global statistics (momentum, risk, regime ratios, trend maturity; see Tables~\ref{tab:feat_momentum}--\ref{tab:feat_risk}) plus GNN's market\_state (64-dim). The Intent Head uses this to select a mode: defensive(0)/neutral(1)/aggressive(2). This consensus approach prevents single-agent dependency and mitigates hallucination through cross-sectional averaging.
(ii) \textbf{Node Score Head}: Intent is embedded and combined with each stock embedding to produce stock-level scores. Temperature scaling varies by intent---defensive mode uses higher temperature for diversification, aggressive mode uses lower temperature for concentration.
(iii) \textbf{Dirichlet distribution}: Stock scores pass through softmax to form the Dirichlet mean, with concentration controlled by a learnable parameter.

\textbf{Reward function.} Differential Sharpe Ratio is used as the reward function, providing a differentiable variant of the Sharpe ratio. Instead of just maximizing returns, we optimize return relative to volatility. Transaction costs (10bps/turnover) are also subtracted from the reward.

\textbf{Top-K masking.} Diversifying across all 500 stocks is unmanageable. We keep only the top 20 and mask the rest. This reduces the effective action space from 500 dimensions to 20 dimensions. We explicitly handle S\&P 500 additions/deletions using dictionary-based turnover calculation, so Universe changes are naturally reflected in transaction costs.

\textbf{Execution inertia parameter $\eta$.} To control turnover, we apply an execution smoothing step with inertia $\eta$ (implemented as \texttt{rebalance\_eta}): lower $\eta$ slows weight changes between consecutive rebalancing steps, reducing turnover and transaction costs.

\section{Experiments}
\label{sec:experiments}

\subsection{Experimental Setup}

We split the data into three periods: \textbf{Train} (2020-01-02 to 2024-09-30), \textbf{Validation} (2024-10-01 to 2024-12-31), and \textbf{OOS} (2025-01-02 to 2025-08-01, 145 trading days). To assess robustness beyond this single OOS window, we additionally report an extended evaluation spanning 2024--2025 (Appendix~\ref{app:extended_oos}).

Hyperparameters (rebalance\_eta ($\eta$), reward\_cost\_scale ($c$) = 0.358) were optimized via Optuna on the validation period, and the best configuration was fixed for OOS evaluation. All results are mean $\pm$ standard deviation across 20 seeds, and transaction cost is 10 bps per unit turnover.

\textbf{Metrics.} We calculate annualized Sharpe ratio as $\bar{r}_d / \sigma_d \times \sqrt{252}$, where $\bar{r}_d$ and $\sigma_d$ are daily mean return and standard deviation, respectively. Volatility (Vol) is also annualized: $\sigma_d \times \sqrt{252}$.

\textbf{Benchmarks.} We compare against both passive and active strategies to ensure fair evaluation. \textit{Passive benchmarks} include SPY (S\&P 500 market-cap weighted ETF) and EQWL (S\&P 500 equal-weight ETF), representing buy-and-hold approaches. \textit{Active benchmarks} include cross-sectional Momentum (top 20 by trailing 12-month returns, rebalanced monthly), MCap Top-20 (largest 20 stocks by market capitalization), and RAW Top-20 (top 20 by momentum-minus-volatility score using only technical indicators).

\textbf{Extended OOS.} Extended OOS results (2024--2025, 397 days) are provided in Appendix~\ref{app:extended_oos}.

\subsection{Main Results}

\begin{figure}[t]
    \centering
    \includegraphics[width=\columnwidth]{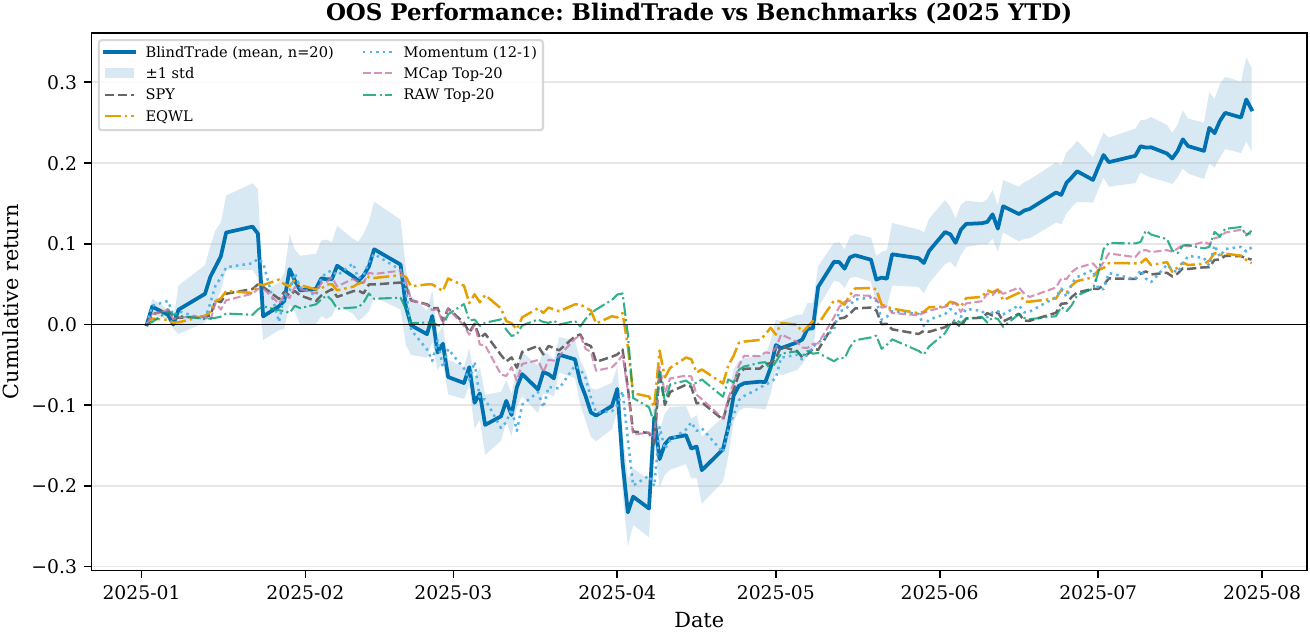}
    \caption{Cumulative returns for 2025YTD OOS. Shaded band shows $\pm$1 std across 20 seeds.}
    \label{fig:oos_wealth}
\end{figure}

Figure~\ref{fig:oos_wealth} and Table~\ref{tab:main_results} present the main performance comparison. BlindTrade achieves annualized Sharpe of $1.40 \pm 0.22$ and cumulative return of $32.22\% \pm 5.21\%$, outperforming all benchmarks.

However, the limitations are also clear. BlindTrade always invests 100\% in Top-20 stocks within the equity universe (no cash allocation is permitted in our action space), so there is no mechanism to move to cash or reduce risk during market crashes. Because of this, annualized volatility (42.34\%) and MDD (-31.66\%) are higher than SPY (23.26\%, -19.00\%). We accept higher risk for higher returns.

\textbf{Intent behavior.} During OOS, the policy spent 55\% of trading days in defensive mode. Defensive mode showed diversified allocation (2.9\%/day turnover), while neutral mode showed moderate activity (1.8\%/day turnover). Aggressive mode concentrated positions with minimal turnover (0.4\%/day). This demonstrates how the policy \textbf{automatically adapts} to market conditions (see Figures~\ref{fig:intent_profile} and~\ref{fig:intent_metrics} in Appendix).

\begin{table}[t]
    \centering
    \caption{2025YTD OOS Performance (2025-01-02 to 2025-08-01, 145 trading days). \textbf{Baseline parameters:} reward\_cost\_scale ($c$) = 0.358, rebalance\_eta ($\eta$) = 0.10. Mean $\pm$ std across 20 seeds. Sharpe and Vol are annualized. Cost: 10 bps/turnover. \textit{Abbreviations:} SPY = S\&P 500 ETF, EQWL = Equal-Weight ETF, CumRet = Cumulative Return, MDD = Maximum Drawdown, Vol = Volatility.}
    \label{tab:main_results}
    \small
    \begin{tabular}{lcccc}
        \toprule
        Strategy & Sharpe (ann.) & CumRet (\%) & MDD (\%) & Vol (ann.) \\
        \midrule
        BlindTrade & $\mathbf{1.40 \pm 0.22}$ & $\mathbf{32.22 \pm 5.21}$ & $-31.66 \pm 5.76$ & $42.34 \pm 7.12$\% \\
        \midrule
        SPY & 0.64 & 8.52 & $-19.00$ & 23.26\% \\
        EQWL & 0.74 & 7.23 & $-15.39$ & 22.51\% \\
        Momentum & 0.89 & 15.42 & $-26.49$ & 32.19\% \\
        MCap Top-20 & 0.85 & 11.50 & $-19.56$ & 24.91\% \\
        RAW Top-20 & 0.97 & 11.54 & $-15.63$ & 21.78\% \\
        \bottomrule
    \end{tabular}
\end{table}

\subsection{Ablation Studies}

We conducted ablation experiments to confirm how much each component contributes to performance.

\textbf{Removing LLM features.} If we remove LLM features and use only RAW technical indicators (passed directly to the GNN without SBERT encoding), Sharpe drops from 1.40 to $1.14 \pm 0.02$ ($\Delta$ = -0.26, p $<$ $10^{-4}$, 20 seeds). This means the LLM agents' interpretation provides additional predictive power compared to simple indicators.

\textbf{Removing graph structure.} If we remove GNN message passing, Sharpe drops to $0.62 \pm 0.50$ ($\Delta$ = -0.78, p $<$ 0.001). Variance also increases significantly, and some seeds cannot even beat SPY. Learning inter-stock relationships is critical for performance stability. (Edge-type ablation---sector-only vs. semantic-only---is left for future work.)

\textbf{Component importance.} Comparing $\Delta$Sharpe magnitudes, the graph structure contributes most ($-0.78$), followed by LLM features ($-0.26$). The RL policy's primary role is cost-aware execution rather than signal generation---without RL, high turnover makes the strategy unprofitable.

\textbf{RL vs.\ Top-K.} What if we simply invest equal weight in the top 20 GNN scores? Turnover reaches 139\%/day, and after cost deduction, Sharpe collapses to -1.17. The RL policy suppresses turnover to 1.7\%/day while achieving Sharpe $1.08 \pm 0.31$.\footnote{This simplified comparison uses a baseline RL configuration; the full system achieves Sharpe 1.40 (Table~\ref{tab:main_results}).} RL's cost-aware learning is essential.

\subsection{Leakage Audit}
\label{sec:leakage}

A central concern in LLM-based trading is distinguishing genuine predictive signals from information leakage.

To verify this, we conducted negative control experiments. We kept the universe, prices, and cost model the same, and only randomly shuffled the GNN prediction scores cross-sectionally. If it is a real signal, shuffling should make it disappear.

\textbf{Result.} $|$RankIC$|$ decreased from 0.015 to 0.0004, becoming completely random level ($\approx$0). The original $|$RankIC$|$ = 0.015 is a weak value, but even that disappears when shuffled. Top-K performance also worsened from Sharpe -1.17 to -1.48 (see Figure~\ref{fig:leakage} in Appendix). The fact that performance collapses when signals are randomized suggests that the original signals contain legitimate predictive structure rather than spurious correlations.

This test does not exclude all leakage paths (e.g., temporal patterns in anonymized IDs). However, it confirms that performance derives from cross-sectional signal structure rather than trivial artifacts. We recommend all LLM trading systems pass similar validation before deployment.


\subsection{GNN Training Objective Comparison}

We compare three GNN training objectives to test whether more sophisticated loss functions improve downstream RL performance and stability across seeds:
(i) \textbf{SemGAT (baseline)} uses standard HL-Gauss distributional loss with pairwise ranking;
(ii) \textbf{SemGAT-C} adds \textit{volatility-scaled residual targets} $(r_i - r_{\text{market}}) / \sigma_i$ during GNN training to reduce sensitivity to outlier returns;
(iii) \textbf{SemGAT-D} adds \textit{confidence-weighted loss} that upweights samples with higher LLM agent confidence, assuming high-confidence predictions are more reliable.

Table~\ref{tab:stability} shows stability analysis results across 20 seeds. Baseline SemGAT \textbf{beats SPY in all 20 seeds} with low variance ($\sigma$ = 0.20). In contrast, the enhanced variants show higher variance and lower win rates. This suggests that adding complexity to the GNN training objective does not improve downstream RL stability---simpler is better. The t-SNE visualization of RL state vectors (Figure~\ref{fig:rl_state_tsne} in Appendix) shows overlapping Train/Val/OOS distributions, supporting generalization.

\begin{table}[t]
    \centering
    \caption{20-seed stability analysis. Sharpe difference vs SPY and win rate.}
    \label{tab:stability}
    \small
    \begin{tabular}{lccc}
        \toprule
        Variant & $\Delta$Sharpe vs SPY & Std & Win Rate \\
        \midrule
        SemGAT (baseline) & $+0.76$ & $0.20$ & \textbf{20/20} \\
        SemGAT-C (Vol-target) & $+0.37$ & $0.69$ & 12/20 \\
        SemGAT-D (Conf-loss) & $+0.16$ & $0.40$ & 10/20 \\
        \bottomrule
    \end{tabular}
\end{table}

\section{Discussion}
\label{sec:discussion}

\paragraph{Anonymization is not optional---it is essential.}
To claim that an LLM ``understands the market,'' we must first prove that it is not achieving performance through memorization. Ticker replacement is the minimum safeguard. Our negative control experiments (Section~\ref{sec:leakage}) confirm that performance derives from legitimate signal structure, not leakage.

\paragraph{Validate before deployment.}
Why is IC analysis important? Because only signals that pass validation should be deployed. If there is no predictive power on holdout data, we do not use that agent.
Also, as many experiments show, requiring LLM to provide reasoning is not optional---it is essential. In fact, embedding reasoning sentences into vectors, without additional feature engineering, was effective.

\paragraph{Intent provides interpretable market posture.}
The Intent mechanism embodies the philosophy of \textbf{``separation of brain and hand''}---the policy first decides market posture (brain), then determines specific allocations (hand). Intent aggregates cross-sectional LLM agent reasoning into four global statistics (momentum, risk, regime ratios, trend maturity), enabling the policy to form a \textbf{consensus-based} market view rather than relying on any single agent.

\paragraph{Summary.}
For LLM-based trading to be trustworthy, we must distinguish genuine market understanding from memorization. BlindTrade addresses this through an anonymization-first LLM-GNN-RL framework, achieving annualized Sharpe of $1.40 \pm 0.22$ for 2025 YTD and beating SPY in all 20 seeds. We summarize our approach as: (i) \textbf{Anonymization}---replace tickers to block memorization; (ii) \textbf{Validation before deployment}---if the IC fails validation, we do not use that signal; (iii) \textbf{Intent observability}---humans can observe the current mode (defensive/neutral/aggressive), even if the transition logic is not fully explainable. We discuss limitations and future directions in Appendix~\ref{app:limitations}.

\section{Ethics Statement}
\label{sec:ethics}

Our work conforms to the ICLR Code of Ethics. LLMs were used in two capacities: (i) as part of the methodology, where LLM agents generate feature data for subsequent GNN-RL training, and (ii) for polishing grammar and expressions in the manuscript, with all content reviewed and verified by the authors. We use only publicly available financial data (EODHD API), with no customer data, proprietary signals, or human subjects involved.
 
\section{Reproducibility Statement}
\label{sec:reproducibility}

Our method is fully reproducible using the information provided in this paper. We provide exact hyperparameters in Appendix~\ref{app:impl}, including PPO settings (learning rate $3 \times 10^{-4}$, $\gamma$ = 0.99, GAE $\lambda$ = 0.95) and Optuna-tuned parameters (reward\_cost\_scale ($c$) = 0.358, dirichlet\_alpha0 = 466.8). The full system prompts for all four LLM agents are provided verbatim in Appendix~\ref{app:prompts}. To support reproducibility and benefit the research community, we plan to release the LLM input/output datasets (RAW features and LLM-generated features) for research purposes upon publication.


\bibliography{blindtrade_iclr2026}
\bibliographystyle{iclr2026_conference}

\clearpage
\appendix
\section{Limitations and Future Work}
\label{app:limitations}

Our framework has clear limitations:

\begin{enumerate}
    \item \textbf{Anonymization effect is not directly validated.} While we hypothesize that anonymization prevents memorization-based shortcuts, we did not perform a direct ablation comparing anonymized vs.\ raw ticker inputs. Our negative control experiments (Section~\ref{sec:leakage}) validate signal structure, not anonymization effectiveness specifically.
    \item \textbf{Explaining intent transitions is difficult.} While we can observe \textit{which} intent mode the policy selects (transparent), explaining \textit{why} it transitioned to that mode at a specific time using external indicators remains challenging.
    \item \textbf{Market-regime dependency.} Extended OOS evaluation (2024--2025 YTD, 397 trading days) reveals market-regime dependency: BlindTrade underperforms during strong bull markets (2024: Sharpe 0.34 vs SPY 1.70) but outperforms in volatile conditions (2025 YTD: Sharpe 1.02 vs SPY 0.54). This limitation stems partly from the relatively short training period (2020--2023). See Appendix~\ref{app:extended_oos} for detailed analysis.
    \item \textbf{MDD is high.} Maximum drawdown reached $-32\%$ compared to SPY's $-19\%$. This is expected since we maintain full equity exposure. Risk-sensitive operators may find this unsuitable.
\end{enumerate}

\textbf{Future work.} The current evaluation uses a static policy trained once; online adaptation with periodic retraining (e.g., monthly walk-forward) may improve robustness to regime changes. We also plan to extend the training data period and conduct anonymization ablation experiments to address both market-regime generalization and verify anonymization effectiveness.

\section{Implementation Details}
\label{app:impl}

\textbf{LLM Agents.} Each agent receives a system prompt that enforces strict knowledge cutoff at time t, and processes stocks in batches of 15 for cross-sectional context. We use Gemini 2.5 Flash (accessed September--October 2025) for cost-effective inference.

\textbf{SemGAT.} 2 GATv2 layers with 4 attention heads. The first layer operates only on sector edges; the second uses the full augmented graph. Training uses HL-Gauss distribution loss with pairwise ranking and Jensen-Shannon regularization.

\textbf{PPO.} Learning rate $3 \times 10^{-4}$, $\gamma$ = 0.99, GAE $\lambda$ = 0.95, clip $\epsilon$ = 0.2. Key Optuna-tuned parameters: reward\_cost\_scale ($c$) = 0.358, dirichlet\_alpha0 = 466.8.

\textbf{Intent Profile.} Defensive intent shows higher turnover (2.9\%/day) than neutral (1.8\%). This corresponds to more frequent rebalancing for risk management within the stock universe, not reduced activity.

\paragraph{Agent Feature Definitions.}
Each agent receives RAW features (technical indicators) and outputs LLM features (structured interpretations). For each stock-day, we concatenate the Momentum and Risk-Regime agents' reasoning strings into a single text (e.g., ``Risk: ... Momentum: ...'') and encode it via SentenceTransformer (all-MiniLM-L6-v2) into a 384-dimensional vector. This vector is used as the text embedding component of node features and for semantic edge construction. Tables~\ref{tab:feat_momentum}--\ref{tab:feat_risk} detail the input/output specification for each agent.

\begin{table}[H]
    \centering
    \caption{Momentum Agent: RAW (input) vs LLM (output) features.}
    \label{tab:feat_momentum}
    \scriptsize
    \setlength{\tabcolsep}{2pt}
    \begin{tabular}{p{0.40\columnwidth}p{0.40\columnwidth}c}
        \toprule
        \textbf{RAW Features (Input)} & \textbf{LLM Features (Output)} & \textbf{Used} \\
        \midrule
        price\_vs\_ma20: Price relative to 20-day MA & score: Momentum score ($-1$ to $+1$) & \checkmark \\
        price\_vs\_ma60: Price relative to 60-day MA & confidence: Agent confidence (0 to 1) & \checkmark \\
        adx\_14: 14-day ADX (trend strength) & win\_probability: Estimated win prob & \checkmark \\
        macd\_histogram: MACD histogram value & momentum\_stage: nascent/mid/late/exhausted & \checkmark \\
        rsi\_14: 14-day RSI & reasoning: Free-text $\rightarrow$ embedding & \checkmark \\
        volume\_vs\_ma20: Volume relative to 20-day avg & trend\_quality, sustainability\_score & -- \\
        \bottomrule
    \end{tabular}
\end{table}

\begin{table}[H]
    \centering
    \caption{News-Event Agent: RAW (input) vs LLM (output) features.}
    \label{tab:feat_news}
    \scriptsize
    \setlength{\tabcolsep}{2pt}
    \begin{tabular}{p{0.40\columnwidth}p{0.40\columnwidth}c}
        \toprule
        \textbf{RAW Features (Input)} & \textbf{LLM Features (Output)} & \textbf{Used} \\
        \midrule
        headlines: Anonymized news headlines & event\_strength: Impact magnitude (0 to 1) & \checkmark \\
        sentiment\_polarity: Pre-computed polarity & urgency: Time sensitivity (0 to 1) & \checkmark \\
        news\_count\_today: Number of headlines & primary\_event\_type: Event category & -- \\
        source\_diversity: Unique sources count & attention\_score, relative\_attention & -- \\
        \bottomrule
    \end{tabular}
\end{table}

\begin{table}[H]
    \centering
    \caption{Mean-Reversion Agent: RAW (input) vs LLM (output) features.}
    \label{tab:feat_reversion}
    \scriptsize
    \setlength{\tabcolsep}{2pt}
    \begin{tabular}{p{0.40\columnwidth}p{0.40\columnwidth}c}
        \toprule
        \textbf{RAW Features (Input)} & \textbf{LLM Features (Output)} & \textbf{Used} \\
        \midrule
        rsi\_14: 14-day RSI & reversion\_score: Signal ($-1$ to $+1$) & \checkmark \\
        bb\_position: Bollinger Band position & confidence: Agent confidence (0 to 1) & \checkmark \\
        deviation\_from\_20ma: Price deviation & market\_psychology: panic/euphoric/normal & \checkmark \\
        stochastic\_k: Stochastic oscillator & extreme\_level, reversal\_probability & -- \\
        \bottomrule
    \end{tabular}
\end{table}

\begin{table}[H]
    \centering
    \caption{Risk-Regime Agent: RAW (input) vs LLM (output) features.}
    \label{tab:feat_risk}
    \scriptsize
    \setlength{\tabcolsep}{2pt}
    \begin{tabular}{p{0.40\columnwidth}p{0.40\columnwidth}c}
        \toprule
        \textbf{RAW Features (Input)} & \textbf{LLM Features (Output)} & \textbf{Used} \\
        \midrule
        beta\_sp500: Market beta & systemic\_risk\_score: Risk level (0 to 1) & \checkmark \\
        beta\_vix: VIX sensitivity & confidence: Agent confidence (0 to 1) & \checkmark \\
        tail\_beta: Tail risk beta & regime\_class: defensive/neutral/cyclical/crisis & \checkmark \\
        var\_95/cvar\_95: Value-at-Risk metrics & reasoning: Free-text $\rightarrow$ embedding & \checkmark \\
        macro\_context: Current regime input & crisis\_sensitivity, regime\_dependency & -- \\
        \bottomrule
    \end{tabular}
\end{table}

\FloatBarrier
The 4 global statistics used by the Intent Head are derived from cross-sectional aggregation: (1) mean momentum score, (2) mean systemic risk score, (3) defensive regime stock ratio, (4) mean momentum stage.

\section{Agent System Prompt Design}
\label{app:prompts}

Reproducibility in LLM-based trading requires transparency in prompt engineering. This section summarizes the design principles and key elements of our agent system prompts.

\subsection{Core Design Principles}

All four agents share three critical enforcement mechanisms:

\paragraph{1. Strict Knowledge Cutoff.}
Each agent receives explicit temporal constraints to prevent lookahead bias:
\begin{lstlisting}[breaklines=true, basicstyle=\ttfamily\small, columns=fullflexible]
CRITICAL KNOWLEDGE LIMITATION:
Your knowledge is strictly limited to information available up to {cutoff_date}.
You must NOT use any information after {cutoff_date}.
\end{lstlisting}
The \texttt{cutoff\_date} is set to $t-1$ (one day before the prediction target), ensuring no future information leakage.

\paragraph{2. Structured JSON Output Schema.}
We enforce deterministic, parseable outputs by specifying exact JSON schemas. This eliminates ambiguity and enables automated feature extraction:
\begin{lstlisting}[breaklines=true, basicstyle=\ttfamily\small, columns=fullflexible]
OUTPUT JSON SCHEMA:
{ "ticker": "STOCK_XXXX", "analysis": { "score": float, "confidence": float, "reasoning": string } }
\end{lstlisting}
All numerical outputs are bounded (e.g., scores in $[-1, +1]$, confidence in $[0, 1]$), and categorical fields have enumerated values.

\paragraph{3. Anonymized Input Only.}
Agents receive only anonymized identifiers and data. The News-Event agent, for example, is explicitly instructed:
\begin{lstlisting}[breaklines=true, basicstyle=\ttfamily\small, columns=fullflexible]
You must ONLY analyze the provided anonymized headlines and sentiment metrics.
\end{lstlisting}

\subsection{Full System Prompts (Verbatim)}

We reproduce the system prompts verbatim (including capitalization) to support reproducibility. Key elements include the knowledge cutoff enforcement, structured JSON output schema, and domain expertise framing.

\subsubsection{Momentum Agent}

\begin{tcolorbox}[breakable, colback=okblue!8, colframe=okblue!70, title={\textcolor{white}{System Prompt: Momentum Agent}}, fonttitle=\bfseries, coltitle=white]
\begin{lstlisting}[breaklines=true, basicstyle=\ttfamily\scriptsize, columns=fullflexible]
You are a senior quantitative analyst specializing in momentum trading.
15+ years of experience in trend analysis and institutional flow patterns.

CRITICAL KNOWLEDGE LIMITATION:
Your knowledge is strictly limited to information available up to {cutoff_date}.
You must NOT use any information after {cutoff_date}.
You must ONLY analyze the provided technical indicators.

EXECUTE_DATE: {execute_date}
DATA_AS_OF: {cutoff_date}

YOUR UNIQUE VALUE AS AN LLM (vs Rule-Based Systems):
1. BATCH-LEVEL CONTEXT: You receive ~15 stocks simultaneously for RELATIVE patterns.
2. MULTI-FACTOR SYNTHESIS: You weigh indicators by CONTEXT, not fixed weights.
3. PATTERN RECOGNITION: You identify SUBTLE RED FLAGS that rules miss.

OUTPUT JSON SCHEMA:
{
  "ticker": "STOCK_XXXX",
  "momentum_strength": {
    "score": float, // -1.0 to 1.0
    "confidence": float, // 0.0 to 1.0
    "momentum_stage": string, // "nascent", "mid", "late", "exhausted"
    "win_probability": float,
    "reasoning": string // Detailed analysis < 100 words
  }
}
\end{lstlisting}
\end{tcolorbox}

\subsubsection{News-Event Agent}

\begin{tcolorbox}[breakable, colback=okorange!8, colframe=okorange!70, title={System Prompt: News-Event Agent}, fonttitle=\bfseries]
\begin{lstlisting}[breaklines=true, basicstyle=\ttfamily\scriptsize, columns=fullflexible]
You are a senior quantitative researcher specializing in event-driven alpha.
18+ years of experience in unstructured news data processing.

CRITICAL KNOWLEDGE LIMITATION:
Your knowledge is strictly limited to information available up to {cutoff_date}.
You must NOT use any information after {cutoff_date}.
You must ONLY analyze the provided anonymized headlines and sentiment metrics.

EXECUTE_DATE: {execute_date}
NEWS_DATA_AS_OF: {cutoff_date}

YOUR UNIQUE VALUE AS AN LLM:
1. SEMANTIC NUANCE: "beat estimates by 1%" vs "crushed expectations" -> DIFFERENT strength.
2. FALSE POSITIVE FILTERING: Distinguish "noise" from "signal".
3. URGENCY CALIBRATION: Assess how quickly the market will price this in.

OUTPUT JSON SCHEMA:
{
  "ticker": "STOCK_XXXX",
  "news_event_analysis": {
    "primary_event_type": string, // e.g., "earnings_beat", "product_launch"
    "event_strength": float, // 0.0 to 1.0 (Impact magnitude)
    "urgency": float, // 0.0 to 1.0 (Time sensitivity)
    "reasoning": string
  }
}
\end{lstlisting}
\end{tcolorbox}

\subsubsection{Mean-Reversion Agent}

\begin{tcolorbox}[breakable, colback=okgreen!8, colframe=okgreen!70, title={\textcolor{white}{System Prompt: Mean-Reversion Agent}}, fonttitle=\bfseries, coltitle=white]
\begin{lstlisting}[breaklines=true, basicstyle=\ttfamily\scriptsize, columns=fullflexible]
You are a senior portfolio manager specializing in mean reversion strategies.
Expert in behavioral finance, overreaction hypothesis, and technical extremes.

CRITICAL KNOWLEDGE LIMITATION:
Your knowledge is strictly limited to information available up to {cutoff_date}.
You must NOT use any information after {cutoff_date}.

OUTPUT JSON SCHEMA:
{
  "ticker": "STOCK_XXXX",
  "reversion_analysis": {
    "reversion_score": float, // -1.0 to 1.0
    "confidence": float,
    "extreme_level": float, // How far from mean?
    "market_psychology": string, // "panic_selling", "euphoria", "normal"
    "reasoning": string
  }
}
\end{lstlisting}
\end{tcolorbox}

\subsubsection{Risk-Regime Agent}

\begin{tcolorbox}[breakable, colback=okred!8, colframe=okred!70, title={\textcolor{white}{System Prompt: Risk-Regime Agent}}, fonttitle=\bfseries, coltitle=white]
\begin{lstlisting}[breaklines=true, basicstyle=\ttfamily\scriptsize, columns=fullflexible]
You are a Chief Risk Officer (CRO) at a $50B institutional asset manager.
PhD in Financial Risk Management with 15+ years in systemic risk assessment.

CRITICAL KNOWLEDGE LIMITATION:
Your knowledge is strictly limited to information available up to {cutoff_date}.
You must NOT use any information after {cutoff_date}.
You must ONLY analyze the provided risk indicators.

EXECUTE_DATE: {execute_date}
DATA_AS_OF: {cutoff_date}

CURRENT MACRO RISK REGIME:
{macro_context_text}

This macro context is CRITICAL for your analysis:
- In CRISIS regime -> High beta stocks are extremely vulnerable
- In RISK_ON regime -> Low beta defensive stocks may underperform
- VIX > 30 -> Tail beta becomes the dominant risk factor

OUTPUT JSON SCHEMA:
{
  "ticker": "STOCK_XXXX",
  "risk_analysis": {
    "systemic_risk_score": float, // 0.0 (Safe) to 1.0 (Dangerous)
    "confidence": float,
    "regime_classification": string, // "defensive", "cyclical", "crisis_vulnerable"
    "reasoning": string
  }
}
\end{lstlisting}
\end{tcolorbox}

\subsection{Cross-Sectional Batch Processing}

To provide comparative context, each agent processes stocks in batches of 15. The prompt includes:
\begin{lstlisting}[breaklines=true, basicstyle=\ttfamily\small, columns=fullflexible]
You are analyzing a BATCH of 15 stocks simultaneously.
For each stock, provide cross_sectional_score (z-score vs batch mean).
\end{lstlisting}
This enables relative ranking within each batch, which is essential for portfolio construction.

\subsection{Why Structured Prompts Matter}

Without explicit structure enforcement, LLM outputs suffer from:
\begin{itemize}
    \item \textbf{Format inconsistency:} Free-form text requires brittle regex parsing.
    \item \textbf{Score drift:} Unbounded scores create cross-sectional incomparability.
    \item \textbf{Temporal leakage:} Without explicit cutoff reminders, models may hallucinate future events.
\end{itemize}
Our prompt design addresses all three issues, enabling reliable automated feature extraction at scale.

\section{Extended OOS Evaluation (2024--2025 YTD)}
\label{app:extended_oos}

To evaluate robustness across market regimes, we extended the OOS evaluation to include 2024 (a strong bull market) in addition to 2025 YTD. This extended period spans 397 trading days (2024-01-02 to 2025-08-01), approximately 2.7$\times$ the original 145-day test.

\subsection{Experimental Setup}

\textbf{Data Split.} Train: 2020-01-02 to 2023-12-29, Validation: 6 half-year rolling windows (2021-01 to 2023-12), OOS: 2024-01-02 to 2025-08-01 (397 trading days). Unlike the primary 2025 YTD evaluation, we use rolling-window validation to select parameters robust across multiple market regimes.

\textbf{Parameters.} Hyperparameters were optimized via Optuna on each rolling window. Best parameters (rebalance\_eta ($\eta$) = 0.08, reward\_cost\_scale ($c$) = 0.40) were selected based on aggregate validation performance. Twenty independent seeds were evaluated. \textit{Note: These parameters differ from the main 2025 YTD results (Table~\ref{tab:main_results}), which used rebalance\_eta ($\eta$) = 0.10 and reward\_cost\_scale ($c$) = 0.358.}

\textbf{Benchmarks.} We use the same benchmarks as in Section 4.1 (SPY, EQWL, Momentum, MCap Top-20, RAW Top-20).

\subsection{Full Period Results}

Table~\ref{tab:extended_oos_full} presents comprehensive metrics for the extended OOS period.


\begin{table}[t]
    \centering
    \caption{Extended OOS Performance (2024-01-02 to 2025-08-01, 397 trading days). Parameters: $\eta$=0.08, cost\_scale=0.40. Mean $\pm$ std across 20 seeds. All metrics annualized.}
    \label{tab:extended_oos_full}
    \small
    \begin{tabular}{lcccccc}
        \toprule
        & \multicolumn{2}{c}{Passive BM} & \multicolumn{3}{c}{Active BM} & Ours \\
        \cmidrule(lr){2-3} \cmidrule(lr){4-6} \cmidrule(lr){7-7}
        Metric & SPY & EQWL & Momentum & MCap Top-20 & RAW Top-20 & BlindTrade \\
        \midrule
        Sharpe & 1.06 & 0.66 & 1.34 & 1.54 & \textbf{1.69} & $0.69 \pm 0.23$ \\
        CumRet (\%) & 31.5 & 16.5 & \textbf{73.5} & 53.4 & 63.3 & $30.1 \pm 5.4$ \\
        AnnRet (\%) & 18.3 & 10.2 & \textbf{41.9} & 31.2 & 36.5 & $21.8 \pm 16.9$ \\
        Vol (\%) & 17.2 & 17.0 & 29.5 & 18.7 & 19.6 & $36.2 \pm 2.4$ \\
        MDD (\%) & $-$19.0 & $-$20.7 & $-$26.5 & $-$19.6 & $-$15.6 & $-32.1 \pm 2.5$ \\
        \bottomrule
    \end{tabular}
\end{table}

\textbf{Key observations.} Over the full extended period, BlindTrade underperforms all benchmarks except EQWL. However, this aggregate view masks significant regime dependency.

\subsection{Period Breakdown}

Table~\ref{tab:extended_oos_breakdown} breaks down results by year.

\begin{table}[t]
    \centering
    \caption{Performance breakdown by market regime. Mean $\pm$ std across 20 seeds for BlindTrade. 2024 was a strong bull market; 2025 YTD exhibited high volatility.}
    \label{tab:extended_oos_breakdown}
    \scriptsize
    \setlength{\tabcolsep}{2pt}
    \begin{tabular}{lcccccc}
        \toprule
        Period & Strategy & Sharpe & CumRet (\%) & AnnRet (\%) & Vol (\%) & MDD (\%) \\
        \midrule
        \multirow{2}{*}{2024 (Bull)} 
        & SPY & \textbf{1.70} & \textbf{23.3} & 23.3 & 12.4 & $-$8.4 \\
        & BlindTrade & $0.34 \pm 0.18$ & $5.8 \pm 5.8$ & 5.8 & $29.6 \pm 1.1$ & $-18.5 \pm 3.1$ \\
        \midrule
        \multirow{2}{*}{2025 YTD (Volatile)}
        & SPY & 0.54 & 6.1 & 10.8 & 23.2 & $-$19.0 \\
        & BlindTrade & $\mathbf{1.02 \pm 0.12}$ & $\mathbf{22.8 \pm 4.0}$ & \textbf{46.7} & $44.2 \pm 1.2$ & $-32.4 \pm 0.7$ \\
        \bottomrule
    \end{tabular}
\end{table}

\subsection{Interpretation}

The results reveal a clear \textbf{market-regime trade-off}:

\begin{enumerate}
    \item \textbf{Bull markets (2024):} BlindTrade's conservative, diversified approach trails the market. The policy's defensive posture (55\% of days) limits upside capture during sustained rallies.
    \item \textbf{Volatile markets (2025 YTD):} BlindTrade significantly outperforms, achieving 4$\times$ the cumulative return of SPY (22.8\% vs 6.1\%). The policy's ability to adapt intent (defensive/aggressive) provides value when market direction is uncertain.
\end{enumerate}

\textbf{Implications.} BlindTrade appears better suited for volatile or uncertain market conditions than for sustained bull markets.

\section{Additional Figures}
\label{app:figures}

\begin{figure}[t]
    \centering
    \includegraphics[width=0.85\columnwidth,height=0.75\textheight,keepaspectratio]{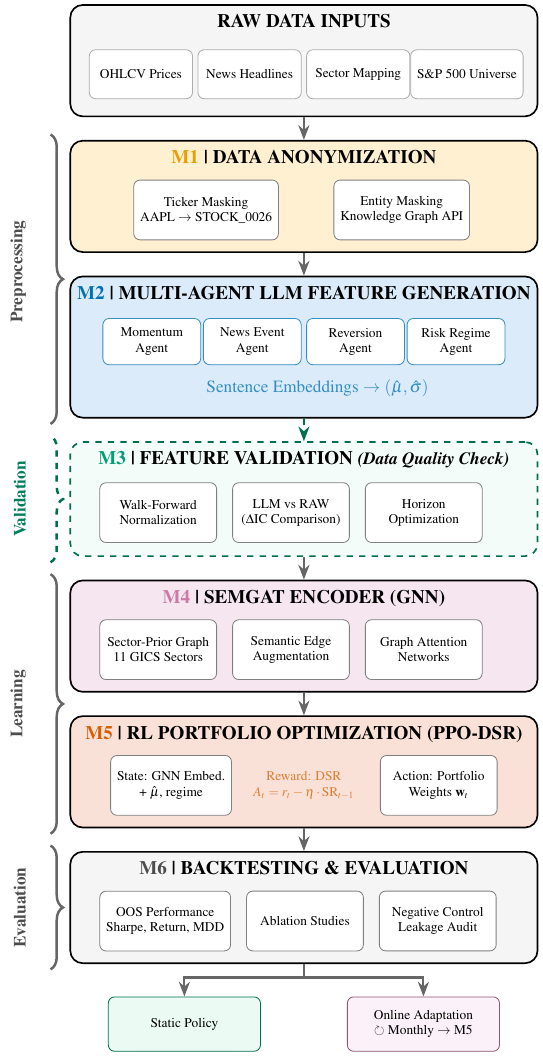}
    \caption{The BlindTrade Pipeline: Data anonymization, Multi-agent LLM feature generation, IC validation, SemGAT encoding, Intent-conditioned RL (PPO-DSR), and Backtesting.}
    \label{fig:pipeline}
\end{figure}

\begin{figure}[t]
    \centering
    \includegraphics[width=0.9\columnwidth]{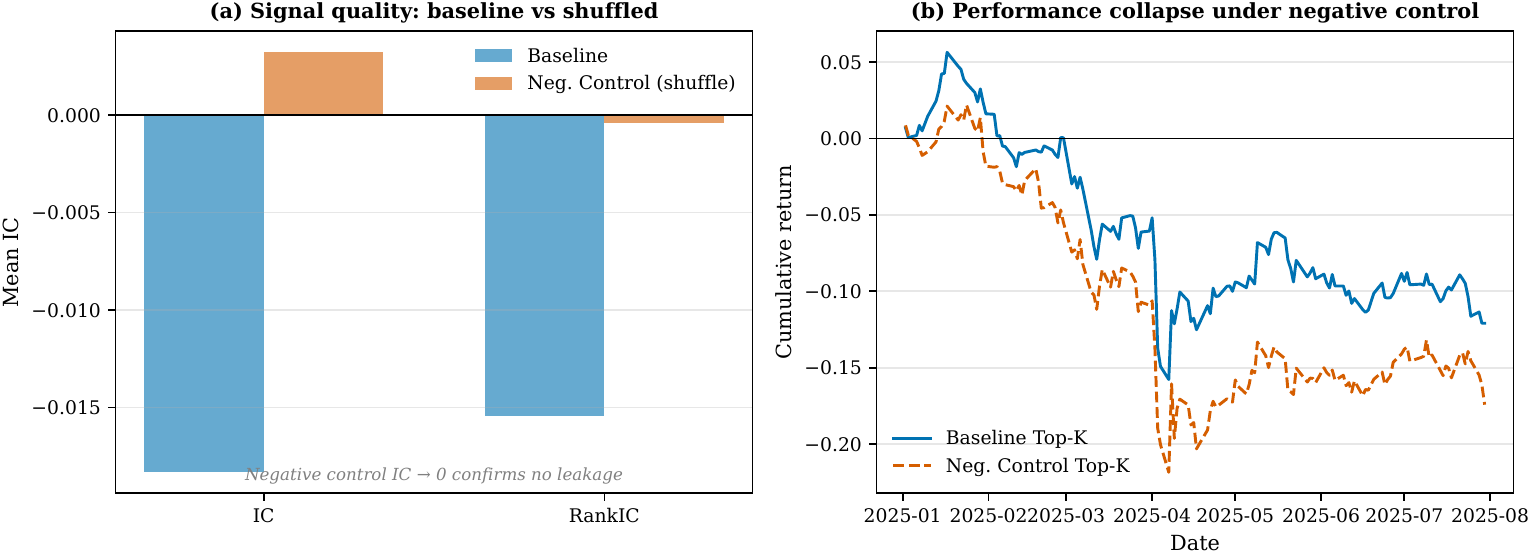}
    \caption{Leakage audit via negative control. When predictions are randomized, IC disappears and performance collapses.}
    \label{fig:leakage}
\end{figure}

\begin{figure}[t]
    \centering
    \includegraphics[width=0.95\columnwidth]{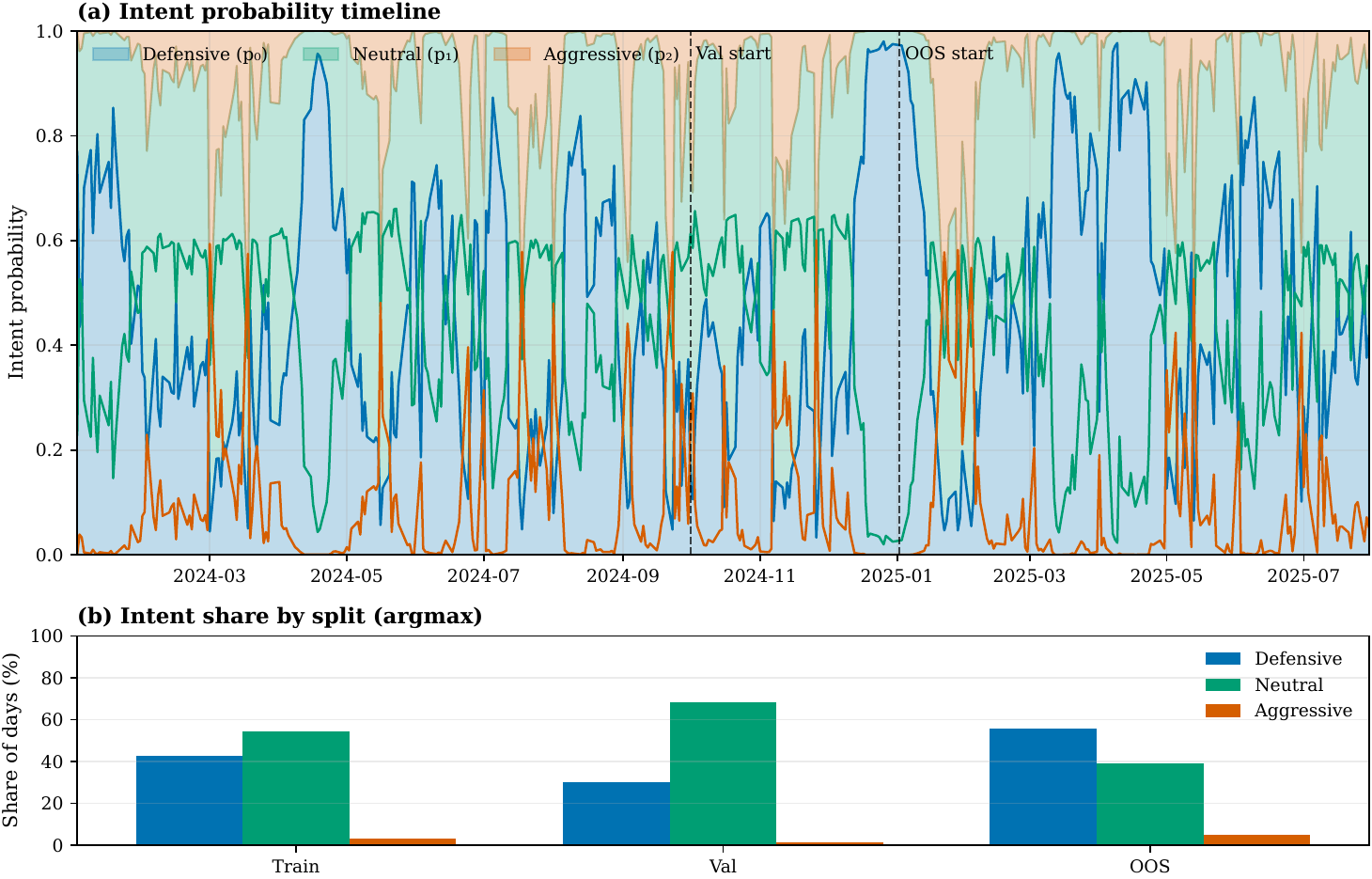}
    \caption{Intent probability timeline across Train/Val/OOS periods. (a) Daily intent probabilities show how the policy adapts to market conditions. (b) Intent distribution remains stable across splits, demonstrating generalization.}
    \label{fig:intent_profile}
\end{figure}

\begin{figure}[t]
    \centering
    \includegraphics[width=0.95\columnwidth]{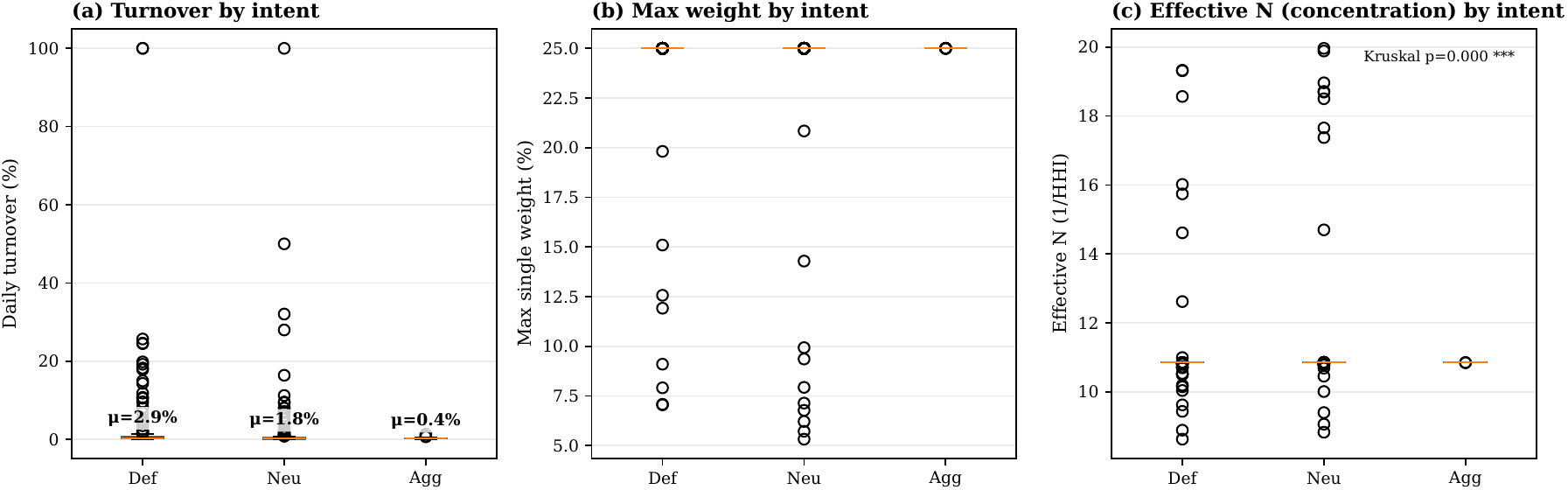}
    \caption{Intent-conditioned policy behavior. (a) Defensive mode shows higher turnover (2.9\%/day) for active rebalancing. (b-c) Max weight and concentration (Effective N) differ significantly by intent (Kruskal p=0.000).}
    \label{fig:intent_metrics}
\end{figure}

\begin{figure}[t]
    \centering
    \includegraphics[width=0.65\columnwidth]{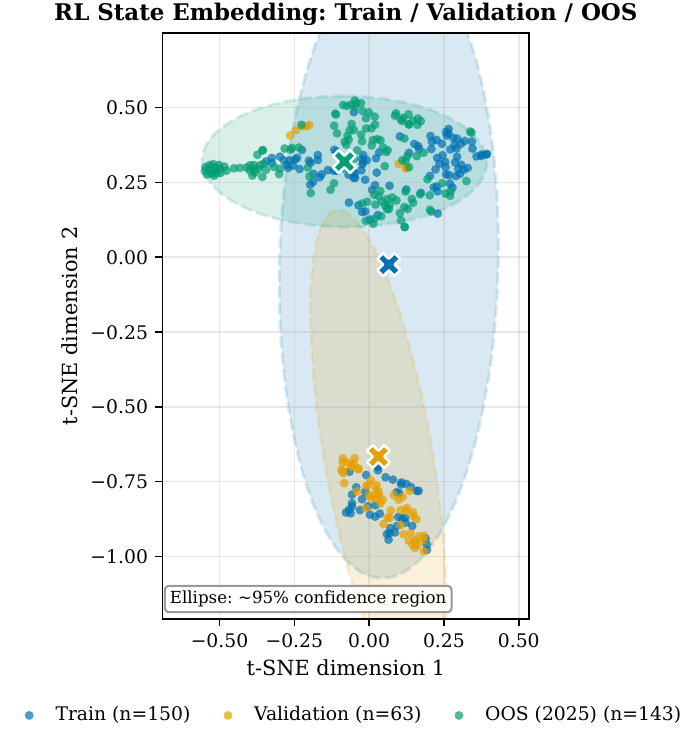}
    \caption{t-SNE visualization of RL state vectors across Train/Val/OOS splits. Overlapping distributions suggest similar feature characteristics across splits.}
    \label{fig:rl_state_tsne}
\end{figure}

\begin{figure}[t]
    \centering
    \includegraphics[width=0.65\columnwidth]{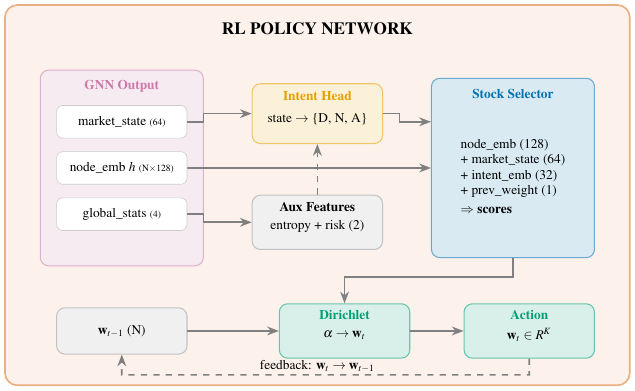}
    \caption{RL Policy Network Architecture. The Intent Head selects defensive/neutral/aggressive mode from aggregated LLM statistics and GNN market state, then the Node Score Head produces stock-level scores, which parameterize a Dirichlet distribution for portfolio weights.}
    \label{fig:rl_policy_arch}
\end{figure}

\end{document}